\documentclass[letterpaper, 10 pt, conference]{ieeeconf}  % Comment this line out if you need a4paper

\IEEEoverridecommandlockouts                              % This command is only needed if 
\usepackage{amsmath}
\usepackage{amsfonts}
\usepackage{dsfont}

\usepackage{eso-pic}

\usepackage{booktabs} % tabels in the book booktabs style
\usepackage{multirow}
\usepackage{graphicx}

\usepackage{comment}
\usepackage{subcaption}
\usepackage[table]{xcolor}
\usepackage{xspace}
\usepackage{array}

\usepackage{hyperref}
\usepackage[capitalise]{cleveref}

\usepackage{tikz}
\usetikzlibrary{external}
\usepackage{pgfplots}
\pgfplotsset{compat=newest}

\usepackage{graphicx}
\usepackage{rotating}
\usepackage{caption}

\usepackage{amssymb}
\usepackage{pifont}
\newcommand{\cmark}{\ding{51}}%
\newcommand{\xmark}{\ding{55}}%

\definecolor{bosch_red}{RGB}{226, 0, 20}
\definecolor{bosch_green}{RGB}{0, 98, 73}

\newcommand{\onedot}{.}%\@addpunct{.}}

\newcommand{\boldparagraph}[1]{\vspace{0.5em}\noindent{\bf #1.}}

\renewcommand{\paragraph}[1]{\boldparagraph{#1}}

\def\ie{\emph{i.e}\onedot}

\def\etal{\emph{et al}\onedot}

\definecolor{bosch_red}{RGB}{226, 0, 20}
\definecolor{bosch_green}{RGB}{0, 98, 73}

\newcommand{\ours}{ALOOD\xspace} % method name

\newcommand\copyrighttext{%
    \footnotesize \copyright\,2025 IEEE. Personal use of this material is permitted.
    Permission from IEEE must be obtained for all other uses, in any current or future
    media, including reprinting/republishing this material for advertising or promotional
    purposes, creating new collective works, for resale or redistribution to servers or
    lists, or reuse of any copyrighted component of this work in other works.%
}

\newcommand\copyrightnotice{%
    \AddToShipoutPictureFG*{%
        \AtPageLowerLeft{%
            \put(\LenToUnit{\dimexpr1in+\oddsidemargin\relax},\LenToUnit{1.0cm}){%
                \fbox{%
                    \parbox{\dimexpr\textwidth-2\fboxsep-2\fboxrule\relax}{%
                        \copyrighttext
                    }%
                }%
            }%
        }%
    }%
}

\begin{document}

\title{\LARGE \bf \ours: Exploiting Language Representations for LiDAR-based Out-of-Distribution Object Detection}

% \author{
%   \quad Michael Kösel$^{1}$
%   \quad Marcel Schreiber$^{2}$ 
%   \quad Michael Ulrich$^{2}$
%   \quad Claudius Gl\"aser$^{2}$ 
%   \quad Klaus Dietmayer$^{1}$\\
%   $^{1}$Ulm University, Ulm \quad $^{2}$Robert Bosch Corporate Research\\
%   {\tt\small \{first.last\}@uni-ulm.de}\\
%   {\tt\small \{first.last\}@de.bosch.com, michael.ulrich2@bosch.com}
% }

\hyphenation{Mi-cro-tech-no-lo-gy}
\author{Michael Kösel$^{1}$, Marcel Schreiber$^{2}$, Michael Ulrich$^{2}$, Claudius Gl\"aser$^{2}$ and Klaus Dietmayer$^{1}$
	%\thanks{The authors are with:}%
	\thanks{$^{1}$Institute of Measurement, Control, and Microtechnology, Ulm University, Germany, {\tt\small \{first.last\}@uni-ulm.de}}% <-this % stops a space
	%\thanks{$^{2}$Robert Bosch GmbH, Corporate Research, 71272 Renningen, Germany, {\tt\small \{first.last\}@de.bosch.com}}%
    \thanks{$^{2}$Robert Bosch GmbH, Corporate Research, 71272 Renningen, Germany, {\tt\small \{first.last\}@de.bosch.com} and {\tt\small michael.ulrich2@bosch.com}}%
    %\thanks{$^{2}$Robert~Bosch~GmbH, ~Corporate~Research, 71272~Renningen, ~Germany, {\tt\small \{marcel.schreiber, claudius.glaeser\}@de.bosch.com} and {\tt\small michael.ulrich2@bosch.com}}%
}

\maketitle
\copyrightnotice
\thispagestyle{empty}
\pagestyle{empty}

\begin{abstract}

LiDAR-based 3D object detection plays a critical role for reliable and safe autonomous driving systems. 
However, existing detectors often produce overly confident predictions for objects not belonging to known categories, posing significant safety risks. 
This is caused by so-called out-of-distribution~(OOD) objects, which were not part of the training data, resulting in incorrect predictions.
To address this challenge, we propose \ours~(\underline{A}ligned \underline{L}iDAR representations for \underline{O}ut-\underline{O}f-\underline{D}istribution Detection), a novel approach that incorporates language representations from a vision-language model~(VLM).
By aligning the object features from the object detector to the feature space of the VLM, we can treat the detection of OOD objects as a zero-shot classification task.
We demonstrate competitive performance on the nuScenes OOD benchmark, establishing a novel approach to OOD object detection in LiDAR using language representations.
%The source code will be released upon publication.
The source code is available at \url{https://github.com/uulm-mrm/mmood3d}.
\end{abstract}    
\section{Introduction}

In recent years, LiDAR sensors in autonomous vehicles have become more important.  
They are often used for 3D object detection because of their ability to detect and classify objects even at large distances accurately.
Despite significant advances in 3D object detection, deploying these detectors in the real world remains a critical challenge.

Most object detectors operate under a closed-world assumption, meaning they are only reliable for input data coming from the distribution they were trained for.
However, in a practical application, the detector may encounter object categories that it has never seen during training.
For example, an object detector that did not include animals during training may misclassify a deer as another class or not detect it at all, leading to potentially dangerous situations.
%These object types that are not part of the training data are commonly referred to as out-of-distribution~(OOD) objects.
%In comparison, in-distribution~(ID) objects are part of the training distribution and are thus detected reliably.
These object types that are not part of the training data are commonly referred to as out-of-distribution~(OOD) objects, whereas object categories that are part of the training distribution are called in-distribution~(ID).

OOD detection has been extensively studied in image classification~\cite{MSP2016, odin2017, maxlogit2019, energy2020, ming2022delving} and has also been extended to 2D~\cite{du2022vos, Wilson_2023_ICCV, kumar2023normalizing, wu2023deep} and 3D object detection~\cite{huang2022out, koesel2024, soum2024open}.
In the context of OOD detection in 3D object detection, Kösel~\etal~\cite{koesel2024} introduced a post-hoc method that extends a fixed CenterPoint~\cite{yin2021center} with a learnable multilayer perceptron (MLP) that uses random scaling to synthesize OOD objects from ID objects.
While this method showed promising results, it relies heavily on the training distribution and struggles when OOD objects differ significantly from ID objects.
Soum-Fontez~\etal~\cite{soum2024open} build on the work of \cite{koesel2024} and autolabel high-confidence false positive~(FP) detections as OOD.
However, this assumes that there are OOD objects in the training set and that the detector is able to localize them.

%Unlike 2D images, LiDAR data lacks texture and color, making it more difficult to distinguish between classes.  
%This increases the difficulty of OOD detection in LiDAR-based object detection, and motivates approaches that can generalize beyond the ID distribution without direct supervision.  

Vision-language models~(VLMs), such as CLIP~\cite{radford2021learning}, have demonstrated strong generalization capabilities in zero-shot classification tasks.
They achieve this by learning a joint feature space between images and text, allowing models to generalize to novel categories given only textual descriptions.
While VLMs were primarily developed for image-based tasks, recent works have shown their effectiveness for LiDAR data in open-vocabulary object detection~\cite{najibi2023unsupervised} and retrieval~\cite{hess2024lidarclip}.
Building on these insights, we show that language representations from VLMs can also be leveraged for LiDAR-based OOD detection.
\begin{figure}[t!]
  \includegraphics[width=0.48\textwidth]{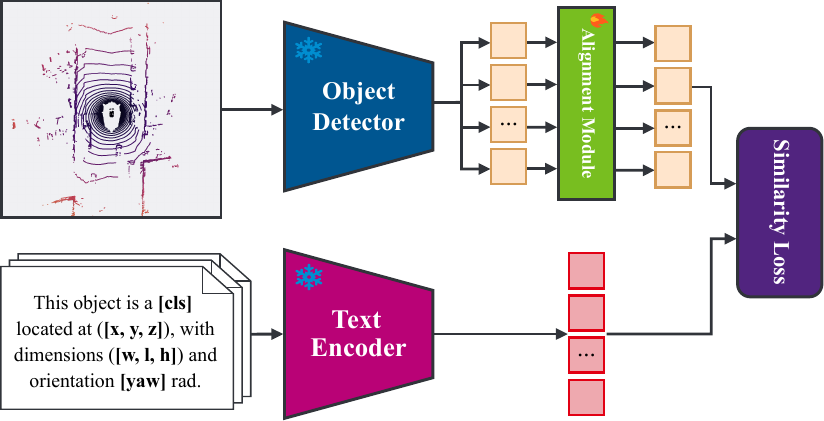}
  \caption{The object features of the LiDAR object detector are aligned to match the embeddings of a frozen text encoder. This allows us to use the zero-shot classification capabilities of VLMs to perform OOD detection.}
  \label{fig:teaser}
\end{figure}

In this work, we propose \ours~(\underline{A}ligned \underline{L}iDAR representations for \underline{O}ut-\underline{O}f-\underline{D}istribution Detection), a novel method, which aligns object features extracted from a 3D LiDAR detector to the CLIP text embedding space, allowing for zero-shot classification of OOD objects.
Unlike other works, we only use CLIP's text encoder and do not need the image encoder during the modality alignment process.
For each predicted object, we generate a prompt as shown in \cref{fig:teaser} and feed it into a frozen text encoder.
The alignment network is then trained to align  the extracted features to the corresponding text embeddings.
This aligned feature space allows us to define OOD detection as a similarity-based classification problem without requiring any OOD supervision.
This means that during inference, the extracted features are compared to each ID text embedding and samples are classified into ID and OOD based on their similarity scores.
The ID text embeddings can be precomputed in an offline step, meaning that the text encoder is not required during inference.
Since our method is also post-hoc, as in ~\cite{koesel2024}, it has no impact on the base detector's performance and requires only training some small additional layers. 
It is trained exclusively on ID data, simplifying the training process.
Our method can detect unknown object categories at test time by exploiting the semantic structure of the CLIP embedding space.

We validate our approach through extensive experiments on the nuScenes OOD benchmark, where we demonstrate better than state-of-the-art performance or competitive results.
Our main contributions can be summarized as follows:  
\begin{itemize}
    \item We propose a novel method for 3D OOD detection by aligning LiDAR object features with language embeddings from CLIP.
    \item Our method achieves state-of-the-art or on-par performance on the nuScenes OOD benchmark without requiring OOD training data. 
    \item We perform detailed ablation studies to understand the impact of our design choices on the OOD detection performance.  
\end{itemize}

\section{Related Work}
\subsection{Out-of-Distribution Detection in Classification}
Ensuring the safe deployment of deep learning-based models in real-world conditions requires reliable OOD detection, which aims to distinguish OOD data from known ID data.
Various approaches have been proposed to tackle this challenge.
Many methods analyze the model output space, \ie, the confidence scores to detect OOD samples~\cite{MSP2016, odin2017, maxlogit2019, energy2020}.
Other approaches rely on the gradient space, \ie, the gradient of OOD samples~\cite{huang2021importance, sun2022dice}
or the activation space, \ie, analyzing the activation of OOD inputs~\cite{lee2018simple, sun2021react, dong2022neural}.
Another line of research utilizes outlier exposure by incorporating an auxiliary OOD dataset~\cite{hendrycks2018ood, yu2019unsupervised}.
Auto-encoders have also been explored for OOD detection~\cite{zhou2022rethinking, hornauer2023heatmap}.
Meanwhile, MCM~\cite{ming2022delving} shows that the zero-shot classification capabilities of CLIP can be used directly for OOD detection.
Similarly, Park~\etal~\cite{park2023powerfulness} employ textual outliers from a CLIP text encoder to perform outlier exposure.

\subsection{OOD Detection in Object Detection}
The concept of OOD detection has also been extended to image-based object detection~\cite{du2022vos, du2022unknown, li2022out}.
VOS~\cite{du2022vos} generates synthetic outlier features by treating the feature space as class-conditional normal distributions and sampling from the low-likelihood regions.
In contrast, SAFE~\cite{Wilson_2023_ICCV} identifies sensitivity-aware layers and trains a post-hoc classifier using features derived from both original and perturbed inputs.
Building on VOS, FFS~\cite{kumar2023normalizing} employs normalizing flows to synthesize outlier features.
Wu~\textit{et al.}~\cite{wu2023deep} introduce feature deblurring diffusion to generate OOD features close to the decision boundary of ID objects. 
In another work, Wu~\textit{et al.}~\cite{wu2024pca} use Principal Component Analysis~(PCA) to extract representative OOD data for training.

\subsection{OOD Detection in 3D Object Detection}
Recently, OOD detection has also been explored in the context of LiDAR-based 3D object detection.
Huang~\etal~\cite{huang2022out} extract features from the PointPillars~\cite{lang2019pointpillars} object detector and evaluate different methods, \ie, Mahalanobis distance~\cite{lee2018simple}, OC-SVM~\cite{scholkopf1999support}, and normalizing flows~\cite{dinh2016density}.
LS-VOS~\cite{piroli2023ls} adopts an approach similar to VOS but employs an auto-encoder to generate outlier features.
Kösel~\etal~\cite{koesel2024} introduce a method that generates synthetic OOD objects by randomly scaling ID objects.
Using these synthetic objects, it is possible to train a simple classifier in a supervised way.
Recently, Soum-Fontez~\etal~\cite{soum2024open} integrated an autolabeling approach into the OOD framework, labeling high-confidence false positive objects as OOD.
Unlike the above methods, our approach avoids any need for synthetic or real OOD data by directly leveraging the knowledge-rich feature space of a VLM.

\subsection{Vision-Language Models for Autonomous Driving}
CLIP~\cite{radford2021learning} learns a joint embedding space for image and text through contrastive training on large amounts of image-text pairs.
Its zero-shot capabilities have enabled broad success in tasks such as open-vocabulary classification and object detection~\cite{gu2022openvocabulary}. Recent work such as VLP~\cite{pan2024vlp} uses VLMs to improve the robustness of planning in autonomous driving by leveraging the reasoning capabilities of language models.
UP-VL~\cite{najibi2023unsupervised} distills image features to the LiDAR features and performs zero-shot classification for the use case of open-vocabulary object detection.
While VLMs have been widely studied in 2D vision, their use in LiDAR-based perception is still underexplored.
In LidarCLIP~\cite{hess2024lidarclip}, features of a LiDAR encoder are aligned with the features of the frozen CLIP image- and text encoder.
This then allows to query point clouds of potentially safety-critical scenes using natural language.
However, most existing methods also require both the image encoder and the text encoder, whereas our method only requires the text encoder.
To the best of our knowledge, we are the first to leverage CLIP’s language embeddings for LiDAR-based OOD object detection.

\section{Method}
\begin{figure*}[t]
\centering
  \includegraphics[width=1\textwidth]{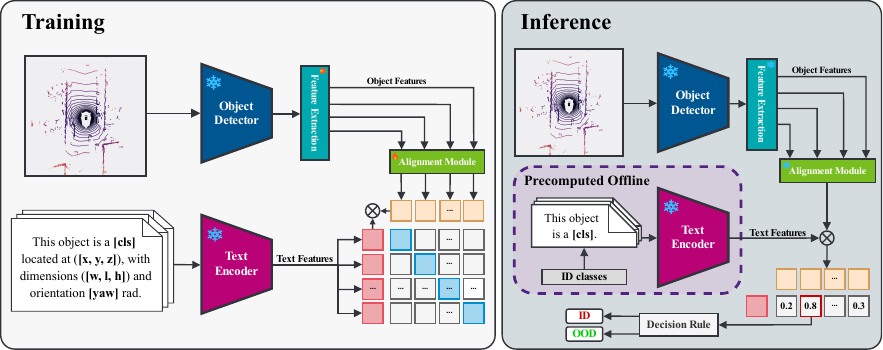}
  \caption{\textbf{Overview of our proposed \ours framework.} Given a frozen LiDAR object detector, we extract features. These object-specific features are aligned to text features from a language model by generating text descriptions for each object. During inference, these aligned object features are compared to cached ID text embeddings using cosine-similarity. }
  \label{fig:overview}
\end{figure*}
\ours introduces a novel OOD detection method that takes advantage of the capabilities of language models. 
We extract features from a pre-trained and frozen object detector.
These object features are then aligned to share the VLM feature space, enabling using the VLM's zero-shot classification capabilities.
The text features from the text encoder of the VLM can be precomputed offline, therefore not requiring the VLM during inference.
A detailed overview of our proposed method is given in \cref{fig:overview}.

%\subsection{Problem Formulation}
\subsection{Preliminaries}

%\boldparagraph{OOD Object Detection}
We address the task of post-hoc OOD detection in LiDAR-based object detection.
Given a point cloud $p \in \mathbb{R}^{N \times 4}$, where each point consists of $(x, y, z, intensity)$, a 3D object detector predicts a set of $M$ 3D bounding boxes $B = \{\mathbf{b}_j\}_{j=1}^M,$ with $\mathbf{b}_j = (x_c, y_c, z_c, l, w, h, \theta) \in \mathbb{R}^7$.
The object detector is trained on a set of $K$ ID classes.
At test time, it may encounter unknown OOD objects not seen during training.
Our goal is to enable the detector to distinguish between ID and OOD objects during inference, without retraining the base object detector on OOD data.
We adopt CenterPoint~\cite{yin2021center} as our base detector, which remains frozen throughout our method.

\subsection{Feature Extraction}
\label{sec:feature_extraction}

Let $\mathbf{F} \in \mathbb{R}^{C \times H \times W}$ denote the BEV feature map of the pretrained object detector.
As the base object detector did not encounter OOD objects during training, its features might not be representative for the OOD detection task.
To address this, we apply a lightweight CNN to the last feature map of the detector's encoder, \ie, the neck feature map.  
This CNN comprises two $3 \times 3$ convolution layers with ReLU activation and batch-norm, followed by a residual connection around them.
This design is motivated by SAFE~\cite{Wilson_2023_ICCV}, which finds that residual and batch-norm layers are most sensitive to OOD inputs.
By applying the CNN, we are able to adapt the otherwise frozen feature map, giving us the improved feature map $\mathbf{F}' \in \mathbb{R}^{C \times H \times W}$.
For each detection $\mathbf{b}_j$, we extract a local feature vector $\mathbf{f}_j \in \mathbb{R}^C$ from $\mathbf{F}'$ using center pooling at its center location, following~\cite{koesel2024}.
During training, ground-truth bounding boxes are used instead of the predicted boxes.
In order to incorporate global scene context, we extract a single feature from $\mathbf{F}'$ using adaptive max-pooling across the entire feature map:
\begin{align}
	\mathbf{f}_{\text{scene}} = \text{AdaptiveMaxPool}(\mathbf{F}') \in \mathbb{R}^C.
    \vspace{-1pt}
\end{align}
This global feature vector provides information about the current scene, giving more context for aligning the features.
We combine the local object features $\mathbf{f}_j$ with the global context feature vector $\mathbf{f}_{\text{scene}}$ as follows:  
\begin{align}
	\tilde{\mathbf{f}}_j = (1 - \lambda) \cdot \mathbf{f}_j + \lambda \cdot \mathbf{f}_{\text{scene}} \in \mathbb{R}^C,
    \vspace{-1pt}
\end{align}
where $\lambda \in [0, 1]$ controls the contribution of the scene context feature.
Additionally, object bounding boxes provide critical geometric cues, giving additional information about the object.
We also include this information to aid alignment and subsequent classification.
First, we project the raw bounding box parameters into a $64$-dimensional space to enhance their expressiveness:
\begin{equation}
\label{eq:box}
\vspace{-1pt}
    \mathbf{g}_j = \text{Linear}(\mathbf{b}_j) \in \mathbb{R}^{64}.
\end{equation}
The final object-level feature is obtained by concatenating the enhanced object features and the encoded box features:  
\begin{align}
\mathbf{u}_j = \text{Concat}(\tilde{\mathbf{f}}_j, \mathbf{g}_j) \in \mathbb{R}^{C+64}.
\vspace{-1pt}
\end{align}
These object features are then used in the following for the alignment and classification.

\subsection{Modality Alignment}
\label{sec:alignment}

In order to bridge the modality gap between the object features from the LiDAR object detector and the VLM, we align the object features to the text space.
For this, we need to create a natural language description for each object to get corresponding text features.
To this end, we create a text prompt for each object in the scene.
We use two types of textual prompts:
\begin{itemize}
    \item Simple: ``This object is a \texttt{[cls]}.''
    \item Spatial: ``This object is a \texttt{[cls]} located at \\ (\texttt{[x, y, z]}), with dimensions (\texttt{[w, l, h]}) and orientation \texttt{[yaw]} rad.''
\end{itemize}
Each object is randomly assigned to one of the two prompt types during training to improve generalization.
For example, a pedestrian could be represented by either ``This object is a pedestrian.'' or ``This object is a pedestrian located at (0.84, -16.86, -2.20), with dimensions (0.98m, 0.56m, 1.62m) and orientation -1.84 rad.''.
The text encoder remains frozen to prevent catastrophic forgetting~\cite{mccloskey1989catastrophic}.
However, the spatially rich prompts allow the encoder to output unique features for different object instances, without requiring fine-tuning.

Let $\mathbf{t}_j \in \mathbb{R}^{D}$ denote the text embedding from the text encoder for object $j$, with $D$ being the embedding dimension of CLIP.
We align the object features by first projecting them to the CLIP feature space using a simple linear layer as the alignment module:
\begin{equation}
\label{eq:aligned_obj}
    \mathbf{v}_j = \text{Linear}(\mathbf{u}_j) \in \mathbb{R}^{D}.
    \vspace{-1pt}
\end{equation}
We measure the similarity of the aligned object features $\mathbf{v}_j$ and the corresponding text embedding $\mathbf{t}_j$, by calculating the cosine similarity, which is given by
\begin{align}
\label{eq:cos_sim}
    \text{sim}(\mathbf{v}_j, \mathbf{t}_j) = \frac{\mathbf{v}_j \cdot \mathbf{t}_j}{\|\mathbf{v}_j\| \|\mathbf{t}_j\|}.
    \vspace{-1pt}
\end{align}
The standard CLIP style InfoNCE loss~\cite{oord2018infonce} assumes distinct positives.
However, in our case, multiple objects in a batch can share the same class.
To address this, we adapt this loss accordingly.
Let $y_i \in \{1, \dots, K\}$ denote the class label of the $i$-th object.  
We define the set of positive matches for object instance $i$ as $\mathcal{P}_i = \{ j \mid y_j = y_i \}$.
Then the updated contrastive loss is given by
\begin{align}
    \mathcal{L}_{\text{align}} = - \frac{1}{N} \sum_{i=1}^{N} \frac{1}{|\mathcal{P}_i|} \sum_{j \in \mathcal{P}_i} \log \frac{\exp(\text{sim}(\mathbf{v}_i, \mathbf{t}_j)/\tau)}{\sum_{r=1}^{N} \exp(\text{sim}(\mathbf{v}_i, \mathbf{t}_r)/\tau)},
    \vspace{-1pt}
\end{align}
where $\tau$ is a learnable temperature parameter.
This contrastive alignment process ensures that the object features $\mathbf{v}_j$ are embedded in the same feature space as their corresponding text feature counterparts.
As a result, the object features become more discriminative and semantically meaningful.
This alignment is crucial for our subsequent inference step, where OOD detection relies on computing similarity scores in the shared feature space.

\subsection{Inference}
\label{sec:inference}

During training, we align the object features to be in the same feature space as the CLIP text features.
This allows us to compare each object feature to a set of ID text features.
Consequently, we generate an ID text embedding for each ID class.
For simplicity, we only include the object class as part of the prompt, i.e. ``This object is a \texttt{[cls]}''.
These prompts do not include instance-specific information such as bounding boxes, so the resulting embeddings are generic and reusable.
This allows us to precompute the ID text embeddings once and use them directly for all further comparisons.
As a result, we do not need the text encoder of CLIP during inference.
Since we also include the simple prompt format during training, we can perform zero-shot classification accordingly.
Then we can compute the cosine-similarity of each object feature with the set of ID text embeddings using \cref{eq:cos_sim}:
\begin{equation}
    s_i = \text{sim}(\mathbf{v}_j, \mathbf{t}_i), \quad i = 1, \dots, K.
\end{equation}
In order to obtain an OOD score from the similarity logits, we use the maximum logit:
\begin{align}
    s_{\text{max}} =\max_{i=1,\dots,K} s_i.
    \vspace{-1pt}
\end{align}
The underlying intuition is that, since OOD objects are not aligned with any ID class during training, their features should exhibit low cosine similarity to the precomputed ID text embeddings.
Empirically, we found that the maximum logit alone is insufficient for satisfactory OOD separation results.
To this end, we scale the maximum logit score $s_{max}$ by the norm of the aligned object features:
\begin{align}
    s_{\text{max}} = \|\mathbf{v}_j\| \cdot \max_{i=1,\dots,K} s_i.
    \vspace{-1pt}
\end{align}
This strategy is consistent with techniques used in prior work~\cite{gong2024outofdistributiondetectionprototypicaloutlier}, where feature norm scaling is also applied to improve discriminative performance.
We can then use this score inside a decision function to perform OOD classification:
\begin{equation}
\label{eq:decision_rule}
G_{\delta} = \begin{cases}
\text{ID} & \text{if } s_{\text{max}} \geq \delta \\
\text{OOD} & \text{otherwise}
\end{cases},
\vspace{-1pt}
\end{equation}
where $\delta$ is a threshold determining whether or not an object is classified as ID or OOD.
Usually, the threshold $\delta$ is chosen so that a high amount of ID objects are correctly classified~(e.g. $95\,\%$).

\section{Experiments}

\subsection{Experimental Setup}

\boldparagraph{Datasets}
We evaluate our approach on the widely-used automotive dataset  nuScenes~\cite{caesar2020}.
The dataset consists of 1000 driving scenes split into $700$ training, $150$ validation, and $150$ testing scenes.
We adopt the nuScenes OOD benchmark from \cite{koesel2024} and consider the $9$ \emph{void} classes as OOD.
Consequently, only scenes not containing any OOD objects are used for training.
%The dataset consists of $1000$ scenes recorded in Boston and Singapore.
%To evaluate the OOD object detection performance, we use the nuScenes benchmark setup~\cite{koesel2024}, \ie we consider the \emph{void} classes as OOD.
%For KITTI we treat the $\{ \text{\emph{Van, Truck, Person sitting, Tram, Misc}} \}$ classes as OOD.

\boldparagraph{Metrics}
Following standard practice in OOD detection~\cite{MSP2016, odin2017, maxlogit2019, energy2020, koesel2024}, we report the False Positive Rate of OOD objects when the True Positive Rate of ID objects is $95$\,\%~(FPR-$95$), the Area Under the Receiver Operating Characteristic~(AUROC), and the Area Under the Precision-Recall curve~(AUPR).
To accommodate the class imbalance, the AUPR is also split into two metrics: AUPR-S, where ID objects are treated as the positive class, and AUPR-E, where OOD objects are considered the positive class.

%Different from \cite{koesel2024}, we use $2$\,m instead of $0.5$\,m as the matching criteria to align with the nuScenes definition of TP metrics.

\boldparagraph{Baseline Methods}
We compare our proposed method against common OOD classification methods~\cite{MSP2016, odin2017, maxlogit2019, energy2020}, as well as existing work in LiDAR-based OOD object detection~\cite{koesel2024}. %\cite{huang2022out, koesel2024}.
The classification baselines~\cite{MSP2016, odin2017, maxlogit2019, energy2020} directly use the classification logits of the base object detector.
%For learning-based methods, we report the mean and standard deviation across five runs with different random seeds.

\boldparagraph{Implementation Details}
Our method is implemented in PyTorch~\cite{paszke2019pytorch} using the MMDetection3D~\cite{mmdet3d2020} framework.
We use both the voxel-based and pillar-based variants of the CenterPoint~\cite{yin2021center} detector as our base object detectors.
For training, we use the AdamW~\cite{loshchilov2018decoupled} optimizer with an initial learning rate of $1.5 \cdot 10^{-4}$, which is halved every two epochs.
We apply the standard CenterPoint data augmentations.
We empirically set $\lambda$ to $0.1$, to keep the focus on the regional object features, while still including global context information. 
In all experiments, we keep the base CenterPoint detector frozen.
Training is performed for 5 epochs, with a total batch size of $48$.
All experiments are conducted on 4$\times$NVIDIA A6000 GPUs.
The ViT-B/32 variant of CLIP~\cite{radford2021learning} is used as the frozen text encoder.

\subsection{Comparisons with State-of-the-art}
\begin{table*}[h!]
    \centering
    \tabcolsep=7.9mm
    \caption{OOD detection results on the nuScenes OOD benchmark. Best results are in \textbf{bold}. Second best are \underline{underlined}.}
    \vspace{1mm}
    \begin{tabular}{c|l|cccc}
        \toprule
        \textbf{CenterPoint} & \textbf{Method} & \textbf{FPR-95~$\downarrow$} & \textbf{AUROC~$\uparrow$} & \textbf{AUPR-S~$\uparrow$} & \textbf{AUPR-E~$\uparrow$} \\
        \midrule

        \multirow{8}{*}{\rotatebox[origin=c]{90}{\textbf{Voxel}}}
        & Default Score      & 63.00 & 85.10 & 99.70 & 7.91 \\
        & MSP~\cite{MSP2016} & 44.60 & 88.87 & \underline{99.79} & 13.74 \\
        & ODIN~\cite{odin2017} & 48.63 & 87.82 & 99.77 & 11.05 \\
        & MaxLogit~\cite{maxlogit2019} & 55.85 & 85.89 & 99.71 & 10.02 \\
        & Energy~\cite{energy2020} & 73.80 & 82.15 & 99.64 & 5.27 \\
        %& NormFlows~\cite{huang2022out} & 85.88$\pm$8.0 & 66.98$\pm$13.4 & 99.25$\pm$0.4 & 2.98$\pm$1.1 \\
        & Rescaling~\cite{koesel2024} & \textbf{36.96} & \underline{88.96} & 99.73 & \textbf{24.68} \\
        & \cellcolor{gray!20}\ours~(ours) & \cellcolor{gray!20}\underline{37.26} & \cellcolor{gray!20}\textbf{90.15} & \cellcolor{gray!20}\textbf{99.81} & \cellcolor{gray!20}\underline{21.52} \\

       \toprule

        \multirow{8}{*}{\rotatebox[origin=c]{90}{\textbf{Pillar}}}
        & Default Score      & 70.48 & 85.21 & 99.54 & 10.20 \\
        & MSP~\cite{MSP2016} & \underline{45.99} & \underline{90.44} & \underline{99.71} & \underline{19.26} \\
        & ODIN~\cite{odin2017} & 56.46 & 88.84 & 99.66 & 14.37 \\
        & MaxLogit~\cite{maxlogit2019} & 66.90 & 85.64 & 99.54 & 11.22 \\
        & Energy~\cite{energy2020} & 89.17 & 78.76 & 99.33 & 5.72 \\
        %& NormFlows~\cite{huang2022out} &  &  &  &  \\
        & Rescaling~\cite{koesel2024} & 66.74 & 84.17 & 99.48 & 10.17 \\
        & \cellcolor{gray!20}\ours~(ours) & \cellcolor{gray!20}\textbf{38.78} & \cellcolor{gray!20}\textbf{91.18} & \cellcolor{gray!20}\textbf{99.72} & \cellcolor{gray!20}\textbf{24.66} \\

        \bottomrule
    \end{tabular}
    \label{tab:results_nus}
\end{table*}
In this section, we compare our method \ours to both classification approaches as well as existing LiDAR-based object detection methods.
The results of this comparison for the two different CenterPoint variants are given in \cref{tab:results_nus}.
\begin{comment}
\begin{table*}[tb!]
    \centering
    \tabcolsep=10mm
    %\small
%    \begin{tabular}{@{}l|ccccc@{}}
    %\resizebox{\columnwidth}{!}{ 
    \caption{OOD detection results on the nuScenes OOD benchmark. Best results are in \textbf{bold}. Second best is \underline{underlined}.}
    \begin{tabular}{@{}lccccc@{}}
        \toprule
        \textbf{Method}   & \textbf{FPR-95~$\downarrow$}   & \textbf{AUROC~$\uparrow$} & \textbf{AUPR-S~$\uparrow$} & \textbf{AUPR-E~$\uparrow$} \\ \midrule
        %Default Score      & - & - & - & - \\
        Default Score      & 63.00 & 85.10 & 99.70 & 7.91 \\
        MSP~\cite{MSP2016}      & 44.60 & 88.87 & \underline{99.79} & 13.74 \\
        ODIN~\cite{odin2017}     & 48.63 & 87.82 & 99.77 & 11.05 \\
        MaxLogit~\cite{maxlogit2019} & 55.85 & 85.89 & 99.71 & 10.02 \\
        Energy~\cite{energy2020}   & 73.80 & 82.15 & 99.64 & 5.27 \\
        \midrule
        NormFlows~\cite{huang2022out} & 85.88$\pm$8.0 & 66.98$\pm$13.4 & 99.25$\pm$0.4 & 2.98$\pm$1.1 \\
        Rescaling~\cite{koesel2024} & \textbf{36.96}$\pm$0.2 & \underline{88.96}$\pm$0.1 & 99.73$\pm$0.0 & \textbf{24.68}$\pm$0.7 \\        \midrule
        \ours~(ours) & \underline{37.26} & \textbf{90.15} &\textbf{99.81} & \underline{21.52} \\ \bottomrule
    \end{tabular}
    %}
    \label{tab:results_nus}
\end{table*}
\end{comment}
The rescaling-based method of~\cite{koesel2024} achieves the best FPR-95 and AUPR-E for the voxel-based CenterPoint.
This can be partially attributed to the prevalence of many small OOD objects in the nuScenes validation set, which the rescaling augmentation handles well.  
In contrast, our method achieves the best results in AUROC and AUPR-S.
This highlights that our method is able to maintain ID classification performance across multiple thresholds while still correctly classifying OOD objects.
Notably, despite not requiring any OOD data during training, our method still achieves strong AUPR-E scores, further validating its effectiveness for OOD detection.

Looking at the results of the pillar-based CenterPoint detector, our method greatly outperforms the rescaling-based method.
This confirms that our method can better generalize, without any special tuning.
Overall, our method shows consistent and promising results across both detector variants.

\subsection{Ablation Study}
%In order to assess the influence of each component, we conduct an ablation study.
%All experiments are performed on the nuScenes validation set.
In the following, we conduct ablation studies on the nuScenes validation set using the voxel-based CenterPoint detector to analyze the influence of each component on the overall performance.

\subsubsection{Alignment Model}

In this section, we analyze the influence of the alignment module on the OOD classification performance.
In addition to a linear layer, we compare using an MLP with and without ReLU activation.
The results are given in \cref{tab:alignment_model}.
\begin{table}[tb!]
    \centering
    %\tabcolsep=0.7mms
    %\small
%    \begin{tabular}{@{}l|ccccc@{}}
    %\resizebox{\columnwidth}{!}{
    \setlength{\tabcolsep}{4.8pt}
    \caption{Comparison of different modules for aligning the object features to the text features.}
    \begin{tabular}{@{}l|ccccc@{}}
        \toprule
        \textbf{Alignment Model}   & \textbf{FPR-95~$\downarrow$}   & \textbf{AUROC~$\uparrow$} & AUPR-S~$\uparrow$ & \textbf{AUPR-E~$\uparrow$} \\ \toprule
        Linear & \textbf{37.26} & \textbf{90.15} & \textbf{99.81} & \textbf{21.52} \\  
        MLP~(ReLU) & 44.86 & 88.27 & 99.76 & 15.67 \\
        MLP~(no ReLU) & 42.50 & 89.32 & 99.77 & 17.11 \\ \bottomrule
    \end{tabular}
    %}
    \label{tab:alignment_model}
\end{table}
Both the linear and the linear MLP as alignment module perform best.
In comparison, when non-linearity is included through the use of the ReLU activation function, the results get worse.
This suggests that no non-linearity is needed to align the object features to the CLIP text feature space.
We hypothesize that the lower capacity of the linear layer enables faster and more stable convergence, particularly given the limited amount of data.

\subsubsection{Features}
To adapt the features of the frozen object detector for OOD detection, we add a simple CNN to the neck feature map before performing feature extraction.
To further enhance the information on the extracted object features, we combine them with global scene information and encoded bounding box features.
In the following we highlight how they influence the capacity of the features.
\Cref{tab:obj_feats} shows the results of this analysis.
\begin{table}[tb!]
    \centering
    \setlength{\tabcolsep}{2.5pt}
    \caption{Ablation study on the importance of the features used.}
    \begin{tabular}{@{}ccc|cccc}
    \toprule
    $CNN$ & $Global$ & $Boxes$ & \textbf{FPR-95~$\downarrow$}   & \textbf{AUROC~$\uparrow$} & \textbf{AUPR-S~$\uparrow$} & \textbf{AUPR-E~$\uparrow$} \\
    \midrule
    - & - & - & 46.20 & 87.98 & 99.76 & 12.89 \\
    $\checkmark$ & - & - & 47.72 & 87.83 & 99.77 & 11.64 \\
    $\checkmark$ & $\checkmark$ & - & 46.39 & 88.71 & 99.79 & 13.10 \\
    $\checkmark$ & - & $\checkmark$ & 37.30 & 90.05 & 99.81 & 20.23 \\
    - & $\checkmark$ & $\checkmark$ & 40.18 & 89.09 & 99.78 & 18.22 \\
    $\checkmark$ & $\checkmark$ & $\checkmark$ & \textbf{37.26} & \textbf{90.15} & \textbf{99.81} & \textbf{21.52} \\
    \bottomrule
    \end{tabular}
    \label{tab:obj_feats}
\end{table}
Including the encoded bounding box features significantly improves performance across all metrics.  
Further gains are achieved by including the global scene context features.  
Interestingly, when analyzing the first two rows, we observe that the introduction of CNN alone decreases performance.  
However, in the last two rows, where all features are included, the CNN significantly improves performance.  
We attribute this to the fact that the CNN improves the representation of the global scene and thus produces more informative combined features.

\subsubsection{Spatial Prompts}

In this experiment, we investigate the effect of including spatial information, \ie, the object's bounding box, in the text prompt.
The results are given in ~\cref{tab:spatial_info}.
\begin{table}[t]
    \centering
    %\tabcolsep=0.7mm
    %\small
%    \begin{tabular}{@{}l|ccccc@{}}
    %\resizebox{\columnwidth}{!}{
    \setlength{\tabcolsep}{5.2pt}
    \caption{Ablation study on the use of spatial information in the text prompts.}
    \begin{tabular}{@{}c|ccccc@{}}
        \toprule
        \textbf{Spatial Prompt}   & \textbf{FPR-95~$\downarrow$}   & \textbf{AUROC~$\uparrow$} & \textbf{AUPR-S~$\uparrow$} & \textbf{AUPR-E~$\uparrow$} \\ \midrule
        \xmark & 38.56 & 89.29 & 99.79 & 18.40 \\        
        %\midrule
        \cmark & \textbf{37.26} & \textbf{90.15} & \textbf{99.81} & \textbf{21.52} \\ \bottomrule
    \end{tabular}
    %}
    \label{tab:spatial_info}
\end{table}
The results show that including the bounding box information into the prompt improves the performance.
Without this information, the alignment network lacks exposure to sufficiently diverse data during training.
This aligns with the training process of CLIP, where images are aligned to their corresponding image descriptions~\cite{radford2021learning}.
In our case, each object is described by its class and bounding box.

\subsubsection{Prompt Format}

In the previous section, we analyzed the influence of bounding box information in the prompt.
Here, we analyze the overall format of the prompt itself.
Specifically, we compare the following prompt formats:
\begin{enumerate}
    \item A \texttt{[cls]} \texttt{[spatial]}.
    \item This object is a \texttt{[cls]} \texttt{[spatial]}.
    \item A photo of a \texttt{[cls]} \texttt{[spatial]}.
\end{enumerate}
For brevity, \texttt{[spatial]} denotes the spatial prompt from \cref{sec:alignment}, which remains identical for all prompt formats.
\Cref{tab:prompt_format} shows the results of this study.
\begin{table}[]
    \centering
    \caption{Analysis of different prompt formats on the final performance}
    \setlength{\tabcolsep}{1.7pt}
    \renewcommand{\arraystretch}{1.2} % vertical padding
    \begin{tabular}{@{}l|cccc@{}}
        \toprule
        \textbf{Prompt Format} & \textbf{FPR-95~$\downarrow$} & \textbf{AUROC~$\uparrow$} & \textbf{AUPR-S~$\uparrow$} & \textbf{AUPR-E~$\uparrow$} \\ 
        \midrule
        1) A \texttt{[cls]} & \textbf{37.04} & 89.83 & 99.80 & 19.70 \\
        2) This object is a \texttt{[cls]}  & 37.26 & \textbf{90.15} & \textbf{99.81} & \textbf{21.52} \\
        3) A photo of a \texttt{[cls]}  & 38.01 & 89.62 & 99.80 & 18.90 \\
        \bottomrule
    \end{tabular}
    \label{tab:prompt_format}
\end{table}
Prompt format 1) uses neutral language without specifying the modality.
Interestingly, prompt format 1) outperforms the original CLIP-style prompt 3), despite the fact that the CLIP text encoder was not fine-tuned for this task.
Using prompt format 2), we achieve the best results.

\subsubsection{OOD Scoring Function}

\begin{figure}[]
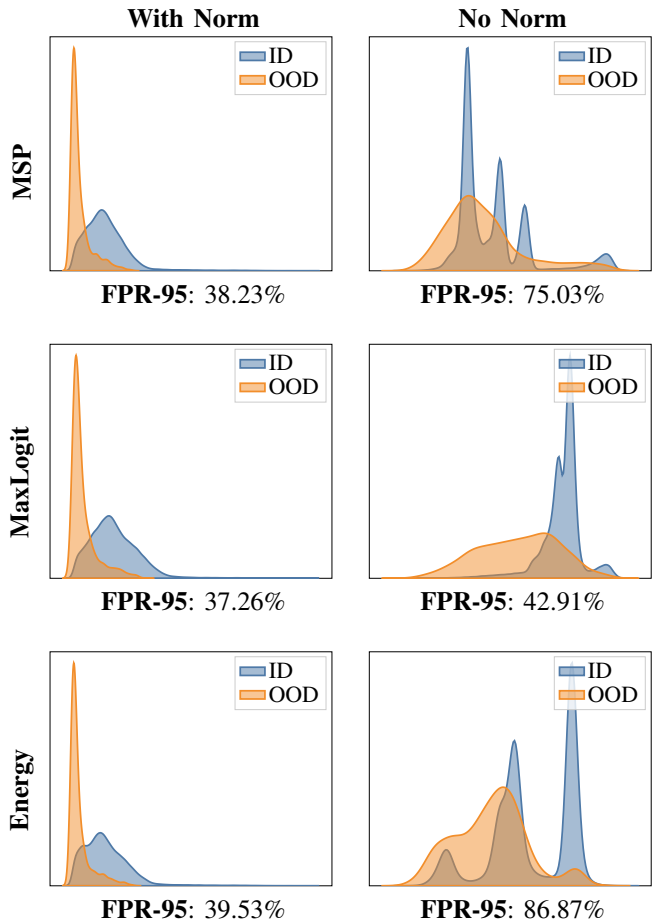

    \centering
    \begin{tabular}{@{}c@{\hskip 0pt}c@{\hskip 0pt}c@{}}
        & \textbf{With Norm} & \textbf{No Norm} \\
        
        \rotatebox[origin=c]{90}{\textbf{MSP}} &
        \begin{tabular}{c}
            \resizebox{0.215\textwidth}{!}{\input{imgs/placeholder/msp_scaled.tex}} \\
            \normalsize \textbf{FPR-95}: 38.23\%
        \end{tabular} &
        \begin{tabular}{c}
            \resizebox{0.215\textwidth}{!}{\input{imgs/placeholder/msp.tex}} \\
            \normalsize \textbf{FPR-95}: 75.03\%
        \end{tabular} \\[3.5ex]
        \\
        \rotatebox[origin=c]{90}{\textbf{MaxLogit}} &
        \begin{tabular}{c}
            \resizebox{0.215\textwidth}{!}{\input{imgs/placeholder/maxlogit_scaled.tex}} \\
            \normalsize \textbf{FPR-95}: 37.26\%
        \end{tabular} &
        \begin{tabular}{c}
            \resizebox{0.215\textwidth}{!}{\input{imgs/placeholder/maxlogit.tex}} \\
            \normalsize \textbf{FPR-95}: 42.91\%
        \end{tabular} \\[3.5ex]
        \\
        \rotatebox[origin=c]{90}{\textbf{Energy}} &
        \begin{tabular}{c}
            \resizebox{0.215\textwidth}{!}{\input{imgs/placeholder/energy_scaled.tex}} \\
            \normalsize \textbf{FPR-95}: 39.53\%
        \end{tabular} &
        \begin{tabular}{c}
            \resizebox{0.215\textwidth}{!}{\input{imgs/placeholder/energy.tex}} \\
            \normalsize \textbf{FPR-95}: 86.87\%
        \end{tabular}
    \end{tabular}
    \caption{Comparison of different OOD score distributions for different scoring methods. Including the object feature norm in the OOD score consistently leads to better results.}
    \label{fig:ood_score_dist}
\end{figure}

In~\cref{fig:ood_score_dist}, we compare different OOD scoring functions.
This function is especially important since it acts as the classifier of the OOD detection task.
Specifically, we compare MaxLogit, MSP, and Energy.
In addition, we also analyze the influence of the object feature norm $\|\mathbf{v}_j\|$ on the results (see \cref{sec:inference}).
By including the norm of the object features in the final OOD score, we get a better separation between the scores for ID and OOD objects.
This can be seen across the board for different score functions, such as MSP or Energy.
This indicates that the feature norm is also a good indicator for OOD objects, since only ID objects are seen during training, leading to different activations for OOD objects.
When including the norm, the score distributions for ID and OOD objects become more distinguishable, with scaled MaxLogit achieving the best overall performance.
In contrast, excluding the norm leads to significant overlap between ID and OOD score distributions, making accurate classification more challenging.
Among all methods, MaxLogit performs best in the norm-free setting, likely because it solely relies on the highest logit value and is independent of other logits.

\section{Conclusion}
In this paper, we propose \ours, a novel approach for OOD object detection in LiDAR that leverages the CLIP feature space for OOD detection.
Our method aligns object features extracted from the CenterPoint object detector with text features from CLIP, bringing them into the unified feature space of CLIP.
Importantly, the CLIP text embeddings for ID classes can be precomputed offline, meaning that the text encoder is not required during inference.
This enables the detection of OOD objects in a zero-shot manner, without requiring OOD samples during training.
%By operating in the feature space of CLIP space, our approach inherits its generalization capabilities and enables more robust OOD classification.
Our experimental results demonstrate that \ours achieves competitive results and even superior results compared to existing methods on the nuScenes OOD benchmark.
These results highlight the potential of VLMs beyond traditional vision tasks, opening up promising directions for future research in LiDAR-based OOD object detection.
In future work, we aim to apply the proposed method to other object detectors, especially two-stage methods.


%%%%%%%%%%%%%%%%%%%%%%%%%%%%%%%%%%%%%%%%%%%%%%%%%%%%%%%%%%%%%%%%%%%%%%%%%%%%%%%%

%%%%%%%%%%%%%%%%%%%%%%%%%%%%%%%%%%%%%%%%%%%%%%%%%%%%%%%%%%%%%%%%%%%%%%%%%%%%%%%%

%%%%%%%%%%%%%%%%%%%%%%%%%%%%%%%%%%%%%%%%%%%%%%%%%%%%%%%%%%%%%%%%%%%%%%%%%%%%%%%%

\bibliographystyle{IEEEtran}
\bibliography{root}

@inproceedings{lang2019pointpillars,
	title        = {{PointPillars: Fast Encoders for Object Detection from Point Clouds}},
	author       = {Lang, Alex H and Vora, Sourabh and Caesar, Holger and Zhou, Lubing and Yang, Jiong and Beijbom, Oscar},
	year         = 2019,
	booktitle    = {Proc. IEEE Conf. Comput. Vis. Pattern Recog.},
	pages        = {12697--12705}
}

@inproceedings{yin2021center,
	title        = {{Center-Based 3D Object Detection and Tracking}},
	author       = {Yin, Tianwei and Zhou, Xingyi and Krahenbuhl, Philipp},
	year         = 2021,
	booktitle    = {Proc. IEEE Conf. Comput. Vis. Pattern Recog.},
	pages        = {11784--11793}
}

@inproceedings{huang2022out,
	title        = {{Out-of-Distribution Detection for LiDAR-Based 3D Object Detection}},
	author       = {Huang, Chengjie and Abdelzad, Vahdat and Mannes, Christopher Gus and Rowe, Luke and Therien, Benjamin and Salay, Rick and Czarnecki, Krzysztof and others},
	year         = 2022,
	booktitle    = {Int. Conf. Intell. Transp. Syst.},
	pages        = {4265--4271},
	organization = {IEEE}
}

@misc{mmdet3d2020,
	title        = {{{MMDetection3D: OpenMMLab}} Next-Generation Platform for General {3D} Object Detection},
	author       = {MMDetection3D Contributors},
	year         = 2020,
	howpublished = {\url{https://github.com/open-mmlab/mmdetection3d}}
}

@inproceedings{dinh2016density,
	title        = {{Density Estimation Using Real NVP}},
	author       = {Dinh, Laurent and Sohl-Dickstein, Jascha and Bengio, Samy},
	year         = 2017,
	booktitle      = {Int. Conf. Learn. Represent.}
}

@article{lee2018simple,
	title        = {{A Simple Unified Framework for Detecting Out-of-Distribution Samples and Adversarial Attacks}},
	author       = {Lee, Kimin and Lee, Kibok and Lee, Honglak and Shin, Jinwoo},
	year         = 2018,
	journal      = {Adv. Neural Inf. Process. Syst.},
	volume       = 31
}

@article{scholkopf1999support,
	title        = {{Support Vector Method for Novelty Detection}},
	author       = {Sch{\"o}lkopf, Bernhard and Williamson, Robert C and Smola, Alex and Shawe-Taylor, John and Platt, John},
	year         = 1999,
	journal      = {Adv. Neural Inf. Process. Syst.},
	volume       = 12
}

@inproceedings{MSP2016,
	title        = {{A Baseline for Detecting Misclassified and Out-of-Distribution Examples in Neural Networks}},
	author       = {Hendrycks, Dan and Gimpel, Kevin},
	year         = 2017,
	booktitle      = {Int. Conf. Learn. Represent.}
}

@inproceedings{maxlogit2019,
  title={{Scaling Out-of-Distribution Detection for Real-World Settings}},
  author={Dan Hendrycks and Steven Basart and Mantas Mazeika and Mohammadreza Mostajabi and Jacob Steinhardt and Dawn Xiaodong Song},
  booktitle={Int. Conf. Mach. Learn.},
  year={2022},
}

@inproceedings{huang2021importance,
	title        = {{On the Importance of Gradients for Detecting Distributional Shifts in the Wild}},
	author       = {Huang, Rui and Geng, Andrew and Li, Yixuan},
	year         = 2021,
	booktitle      = {Advances in Neural Inform. Processing Syst.},
	volume       = 34,
	pages        = {677--689}
}

@article{energy2020,
	title        = {{Energy-Based Out-of-Distribution Detection}},
	author       = {Liu, Weitang and Wang, Xiaoyun and Owens, John and Li, Yixuan},
	year         = 2020,
	journal      = {Adv. Neural Inf. Process. Syst.},
	volume       = 33,
	pages        = {21464--21475}
}

@inproceedings{odin2017,
	title        = {{Enhancing the Reliability of Out-of-Distribution Image Detection in Neural Networks}},
	author       = {Liang, Shiyu and Li, Yixuan and Srikant, R.},
	year         = 2018,
    booktitle      = {Int. Conf. Learn. Represent.}
}

@inproceedings{hendrycks2018ood,
	title        = {{Deep Anomaly Detection with Outlier Exposure}},
	author       = {Hendrycks, Dan and Mazeika, Mantas and Dietterich, Thomas},
	year         = 2019,
	booktitle      = {Int. Conf. Learn. Represent.}
}

@article{sun2021react,
	title        = {{ReAct: Out-of-Distribution Detection with Rectified Activations}},
	author       = {Sun, Yiyou and Guo, Chuan and Li, Yixuan},
	year         = 2021,
	journal      = {Advances in Neural Inform. Processing Syst.},
	volume       = 34,
	pages        = {144--157}
}

@inproceedings{hornauer2023heatmap,
	title        = {{Heatmap-Based Out-of-Distribution Detection}},
	author       = {Hornauer, Julia and Belagiannis, Vasileios},
	year         = 2023,
	booktitle    = {Proc. IEEE/CVF Winter Conf. Appl. Comput. Vis.},
	pages        = {2603--2612}
}

@inproceedings{du2022vos,
	title        = {{VOS: Learning What You Don’t Know by Virtual Outlier Synthesis}},
	author       = {Du, Xuefeng and Wang, Zhaoning and Cai, Mu and Li, Yixuan},
	year         = 2022,
	booktitle      = {Int. Conf. Learn. Represent.}
}

@inproceedings{piroli2023ls,
	title        = {{LS-VOS: Identifying Outliers in 3D Object Detections Using Latent Space Virtual Outlier Synthesis}},
	author       = {Piroli, Aldi and Dallabetta, Vinzenz and Kopp, Johannes and Walessa, Marc and Meissner, Daniel and Dietmayer, Klaus},
    booktitle={Int. Conf. Intell. Transp. Syst.},
    pages={1242--1248},
	year={2023},
}

@inproceedings{caesar2020,
	title        = {{nuScenes: A Multimodal Dataset for Autonomous Driving}},
	booktitle    = {Proc. IEEE Conf. Comput. Vis. Pattern Recog.},
	author       = {Caesar, Holger and Bankiti, Varun and Lang, Alex H. and Vora, Sourabh and Liong, Venice Erin and Xu, Qiang and Krishnan, Anush and Pan, Yu and Baldan, Giancarlo and Beijbom, Oscar},
	year         = {2020},
	pages        = {11621--11631}
}

@inproceedings{Wilson_2023_ICCV,
	title        = {{SAFE: Sensitivity-Aware Features for Out-of-Distribution Object Detection}},
	author       = {Wilson, Samuel and Fischer, Tobias and Dayoub, Feras and Miller, Dimity and S{\"u}nderhauf, Niko},
	year         = 2023,
	month        = {October},
	booktitle    = {Proc. Int. Conf. Comput. Vis.},
	pages        = {23565--23576}
}

@inproceedings{
gu2022openvocabulary,
title={{Open-vocabulary Object Detection via Vision and Language Knowledge Distillation}},
author={Xiuye Gu and Tsung-Yi Lin and Weicheng Kuo and Yin Cui},
booktitle={Int. Conf. Learn. Represent.},
year={2022},
url={https://openreview.net/forum?id=lL3lnMbR4WU}
}

@article{paszke2019pytorch,
	title        = {{PyTorch: An Imperative Style, High-Performance Deep Learning Library}},
	author       = {Paszke, Adam and Gross, Sam and Massa, Francisco and Lerer, Adam and Bradbury, James and Chanan, Gregory and Killeen, Trevor and Lin, Zeming and Gimelshein, Natalia and Antiga, Luca and others},
	year         = 2019,
	journal      = {Adv. Neural Inf. Process. Syst.},
	volume       = 32
}

@inproceedings{wu2023deep,
	title        = {{Deep Feature Deblurring Diffusion for Detecting Out-of-Distribution Objects}},
	author       = {Wu, Aming and Chen, Da and Deng, Cheng},
	year         = 2023,
	booktitle    = {Proc. IEEE Conf. Comput. Vis. Pattern Recog.},
	pages        = {13381--13391}
}

@inproceedings{kumar2023normalizing,
	title        = {{Normalizing Flow Based Feature Synthesis for Outlier-Aware Object Detection}},
	author       = {Kumar, Nishant and {\v{S}}egvi{\'c}, Sini{\v{s}}a and Eslami, Abouzar and Gumhold, Stefan},
	year         = 2023,
	booktitle    = {Proc. IEEE Conf. Comput. Vis. Pattern Recog.},
	pages        = {5156--5165}
}

@inproceedings{radford2021learning,
  title={{Learning Transferable Visual Models From Natural Language Supervision}},
  author={Radford, Alec and Kim, Jong Wook and Hallacy, Chris and Ramesh, Aditya and Goh, Gabriel and Agarwal, Sandhini and Sastry, Girish and Askell, Amanda and Mishkin, Pamela and Clark, Jack and others},
  booktitle={International conference on machine learning},
  pages={8748--8763},
  year={2021},
  organization={PMLR}
}

@INPROCEEDINGS{koesel2024,
  author={Kösel, Michael and Schreiber, Marcel and Ulrich, Michael and Gläser, Claudius and Dietmayer, Klaus},
  booktitle={2024 IEEE Intelligent Vehicles Symposium (IV)}, 
  title={{Revisiting Out-of-Distribution Detection in LiDAR-based 3D Object Detection}}, 
  year={2024},
  volume={},
  number={},
  pages={2806-2813},
}

@article{soum2024open,
  title={{Open-Set 3D object detection in LiDAR data as an Out-of-Distribution problem}},
  author={Soum-Fontez, Louis and Deschaud, Jean-Emmanuel and Goulette, Fran{\c{c}}ois},
  journal={arXiv preprint arXiv:2410.23767},
  year={2024}
}

@article{ming2022delving,
  title={{Delving into Out-of-Distribution Detection with Vision-Language Representations}},
  author={Ming, Yifei and Cai, Ziyang and Gu, Jiuxiang and Sun, Yiyou and Li, Wei and Li, Yixuan},
  journal={Advances in Neural Inform. Processing Syst.},
  volume={35},
  pages={35087--35102},
  year={2022}
}

@article{park2023powerfulness,
  title={{On the Powerfulness of Textual Outlier Exposure for Visual OoD Detection}},
  author={Park, Sangha and Mok, Jisoo and Jung, Dahuin and Lee, Saehyung and Yoon, Sungroh},
  journal={Advances in Neural Inform. Processing Syst.},
  volume={36},
  pages={51675--51687},
  year={2023}
}

@inproceedings{loshchilov2018decoupled,
    title={{Decoupled Weight Decay Regularization}},
    author={Ilya Loshchilov and Frank Hutter},
    booktitle={Int. Conf. Learn. Represent.},
    year={2019},
    url={https://openreview.net/forum?id=Bkg6RiCqY7},
}

@inproceedings{du2022unknown,
  title={{Unknown-Aware Object Detection: Learning What You Don't Know from Videos in the Wild}},
  author={Du, Xuefeng and Wang, Xin and Gozum, Gabriel and Li, Yixuan},
  booktitle={Proc. IEEE Conf. Comput. Vis. Pattern Recog.},
  pages={13678--13688},
  year={2022}
}

@inproceedings{yu2019unsupervised,
  title={Unsupervised out-of-distribution detection by maximum classifier discrepancy},
  author={Yu, Qing and Aizawa, Kiyoharu},
  booktitle={Proc. Int. Conf. Comput. Vis.},
  pages={9518--9526},
  year={2019}
}

@inproceedings{zhou2022rethinking,
  title={{Rethinking Reconstruction Autoencoder-Based Out-of-Distribution Detection}},
  author={Zhou, Yibo},
  booktitle={Proc. IEEE Conf. Comput. Vis. Pattern Recog.},
  pages={7379--7387},
  year={2022}
}

@inproceedings{dong2022neural,
  title={{Neural Mean Discrepancy for Efficient Out-of-Distribution Detection}},
  author={Dong, Xin and Guo, Junfeng and Li, Ang and Ting, Wei-Te and Liu, Cong and Kung, HT},
  booktitle={Proc. IEEE Conf. Comput. Vis. Pattern Recog.},
  pages={19217--19227},
  year={2022}
}

@inproceedings{sun2022dice,
  title={{DICE: Leveraging Sparsification for Out-of-Distribution Detection}},
  author={Sun, Yiyou and Li, Yixuan},
  booktitle={European Conference on Computer Vision},
  pages={691--708},
  year={2022},
  organization={Springer}
}

@inproceedings{li2022out,
  title={{Out-of-Distribution Identification: Let Detector Tell Which I Am Not Sure}},
  author={Li, Ruoqi and Zhang, Chongyang and Zhou, Hao and Shi, Chao and Luo, Yan},
  booktitle={European Conference on Computer Vision},
  pages={638--654},
  year={2022},
  organization={Springer}
}

@inproceedings{pan2024vlp,
  title={{VLP: Vision Language Planning for Autonomous Driving}},
  author={Pan, Chenbin and Yaman, Burhaneddin and Nesti, Tommaso and Mallik, Abhirup and Allievi, Alessandro G and Velipasalar, Senem and Ren, Liu},
  booktitle={Proc. IEEE Conf. Comput. Vis. Pattern Recog.},
  pages={14760--14769},
  year={2024}
}

@incollection{mccloskey1989catastrophic,
  title={{Catastrophic Interference in Connectionist Networks: The Sequential Learning Problem}},
  author={McCloskey, Michael and Cohen, Neal J},
  booktitle={Psychology of learning and motivation},
  volume={24},
  pages={109--165},
  year={1989},
  publisher={Elsevier}
}

@ARTICLE{wu2024pca,
  author={Wu, Aming and Deng, Cheng and Liu, Wei},
  journal={IEEE Trans. Image Process.}, 
  title={{Unsupervised Out-of-Distribution Object Detection via PCA-Driven Dynamic Prototype Enhancement}}, 
  year={2024},
  volume={33},
  number={},
  pages={2431-2446},
}

@inproceedings{hess2024lidarclip,
  title={{LidarCLIP or: How I Learned to Talk to Point Clouds}},
  author={Hess, Georg and Tonderski, Adam and Petersson, Christoffer and {\AA}str{\"o}m, Kalle and Svensson, Lennart},
  booktitle={Proc. IEEE/CVF Winter Conf. Appl. Comput. Vis.},
  pages={7438--7447},
  year={2024}
}

@inproceedings{najibi2023unsupervised,
  title={{Unsupervised 3D Perception with 2D Vision-Language Distillation for Autonomous Driving}},
  author={Najibi, Mahyar and Ji, Jingwei and Zhou, Yin and Qi, Charles R and Yan, Xinchen and Ettinger, Scott and Anguelov, Dragomir},
  booktitle={Proc. Int. Conf. Comput. Vis.},
  pages={8602--8612},
  year={2023}
}

@article{oord2018infonce,
  author       = {A{\"{a}}ron van den Oord and
                  Yazhe Li and
                  Oriol Vinyals},
  title        = {{Representation Learning with Contrastive Predictive Coding}},
  journal      = {CoRR},
  volume       = {abs/1807.03748},
  year         = {2018},
  url          = {http://arxiv.org/abs/1807.03748},
  eprinttype    = {arXiv},
  eprint       = {1807.03748},
}

@misc{gong2024outofdistributiondetectionprototypicaloutlier,
      title={{Out-of-Distribution Detection with Prototypical Outlier Proxy}}, 
      author={Mingrong Gong and Chaoqi Chen and Qingqiang Sun and Yue Wang and Hui Huang},
      year={2024},
      eprint={2412.16884},
      archivePrefix={arXiv},
      primaryClass={cs.CV},
      url={https://arxiv.org/abs/2412.16884}, 
}

\end{document}